\documentclass[lettersize,journal]{IEEEtran}
\usepackage{amsmath,amsfonts}
\usepackage{algorithmic}
\usepackage{array}
\usepackage[caption=false,font=normalsize,labelfont=sf,textfont=sf]{subfig}
\usepackage{textcomp}
\usepackage{stfloats}
\usepackage{url}
\usepackage{verbatim}
\usepackage{graphicx}
\usepackage{cite}
\pdfoutput=1

\usepackage{balance}
\usepackage{xcolor}
\usepackage{soul}
\usepackage{multirow}
\usepackage{adjustbox}
\usepackage[ruled,vlined]{algorithm2e}
\PassOptionsToPackage{ruled,vlined}{algorithm2e}
\usepackage{setspace}
\usepackage{eucal}
\usepackage{booktabs}
\usepackage{pifont}
\usepackage{hyperref}
\usepackage{fancyhdr}
\newcommand{\xmark}{\ding{55}}

\begin{document}

\title{Self-supervised Graphs for Audio Representation Learning with Limited Labeled Data}

\author{Amir Shirian, Krishna Somandepalli~\IEEEmembership{Member,~IEEE}, Tanaya Guha,~\IEEEmembership{Member,~IEEE,}
\thanks{A. Shirian is with the Department of Computer Science, University of Warwick, UK. K. Somandepalli is with Google Research, US. T. Guha is with the School of Computing Science, University of Glasgow, UK.}
}

\markboth{
Journal of Selected Topics in Signal Processing 
~July~2022}%
{Shell \MakeLowercase{\textit{et al.}}: A Sample Article Using IEEEtran.cls for IEEE Journals}


\maketitle

\begin{abstract}
Large-scale databases with high-quality manual labels are scarce in audio domain. We thus explore a self-supervised graph approach to learning audio representations from highly limited labelled data. Considering each audio sample as a graph node, we propose a subgraph-based framework with novel self-supervision tasks to learn effective audio representations. During training, subgraphs are constructed by sampling the entire pool of available training data to exploit the relationship between the labelled and unlabeled audio samples. During inference, we use random edges to alleviate the overhead of graph construction. We evaluate our model on three benchmark audio datasets spanning two tasks: acoustic event classification and speech emotion recognition. We show that our semi-supervised model performs better or on par with fully supervised models and outperforms several competitive existing models. Our model is compact and can produce generalized audio representations robust to different types of signal noise. Our code is available at \href{https://github.com/AmirSh15/SSL_graph_audio}{\texttt{github.com/AmirSh15/SSL\_graph\_audio}}
\end{abstract}


\section{Introduction}
\label{sec:intro}
Large databases with high-quality manual labels are scarce in audio domain. For tasks such as speech-based emotion analysis, manual labels are often difficult to acquire due to the subjectivity involved in the perception and expression of emotion across speakers, language and culture. On the other hand, for tasks such as acoustics event classification, manually labeling a large volume of audio data is simply impractical. Thus a core challenge in audio analysis is to learn from a limited amount of labeled data while taking advantage of larger amount of unlabeled training samples.

\textbf{Why graphs?} Self-Supervised Learning (SSL) has emerged as an effective approach to learning from unlabeled data \cite{doersch2015unsupervised,chen2020big,xie2020self, DevlinCLT19}. We propose an SSL approach on graphs to learn effective audio representations from limited amount of labeled data. Considering each audio sample as a node in a graph, we cast audio classification as a node labeling task. The motivation behind adopting a graph approach is two-fold: (i) It leads to compact models as compared to commonly used recurrent speech models as noted in recent works \cite{shirian2021compact,LiuW21}; (ii) A graph structure, if properly constructed, can efficiently capture the relationship between the small number of available labeled nodes and a larger number of unlabeled nodes. Extensive experiments with standard benchmarks brings out the advantages of graph-based methods in terms of performance compared to the non-graph models.

Following the success of SSL on images, it has been extended to graph data for both fully supervised \cite{tsitsulin2018sgr} and semi-supervised \cite{you2020does,zhu2020self,jin2020self} tasks. Graph SSL tasks usually involve learning the local or global structure, or the context information in the data \cite{jin2020self,tsitsulin2018sgr,you2020does}. Conventional graph tasks such as node clustering and graph partitioning have already been used as SSL tasks \cite{you2020does}. 
In the audio domain, SSL has also started gaining popularity. Several recent papers report that SSL can improve over fully supervised models \cite{tagliasacchi2020pre} while others use crossmodal self-supervision from visual domain \cite{shukla2020does, nagrani2020disentangled}. However, works that use graph approach to learning audio representation is limited. We are aware of only one recent work where an audio sequence has been considered as a line graph to exploit graph signal processing theory to achieve accurate spectral graph convolution \cite{shirian2021compact}.
\begin{figure*}[tb]
\centering
   \includegraphics[width=0.8\linewidth, trim=0mm 0mm 0mm 0mm, clip=true, bb=0 0 531 324]{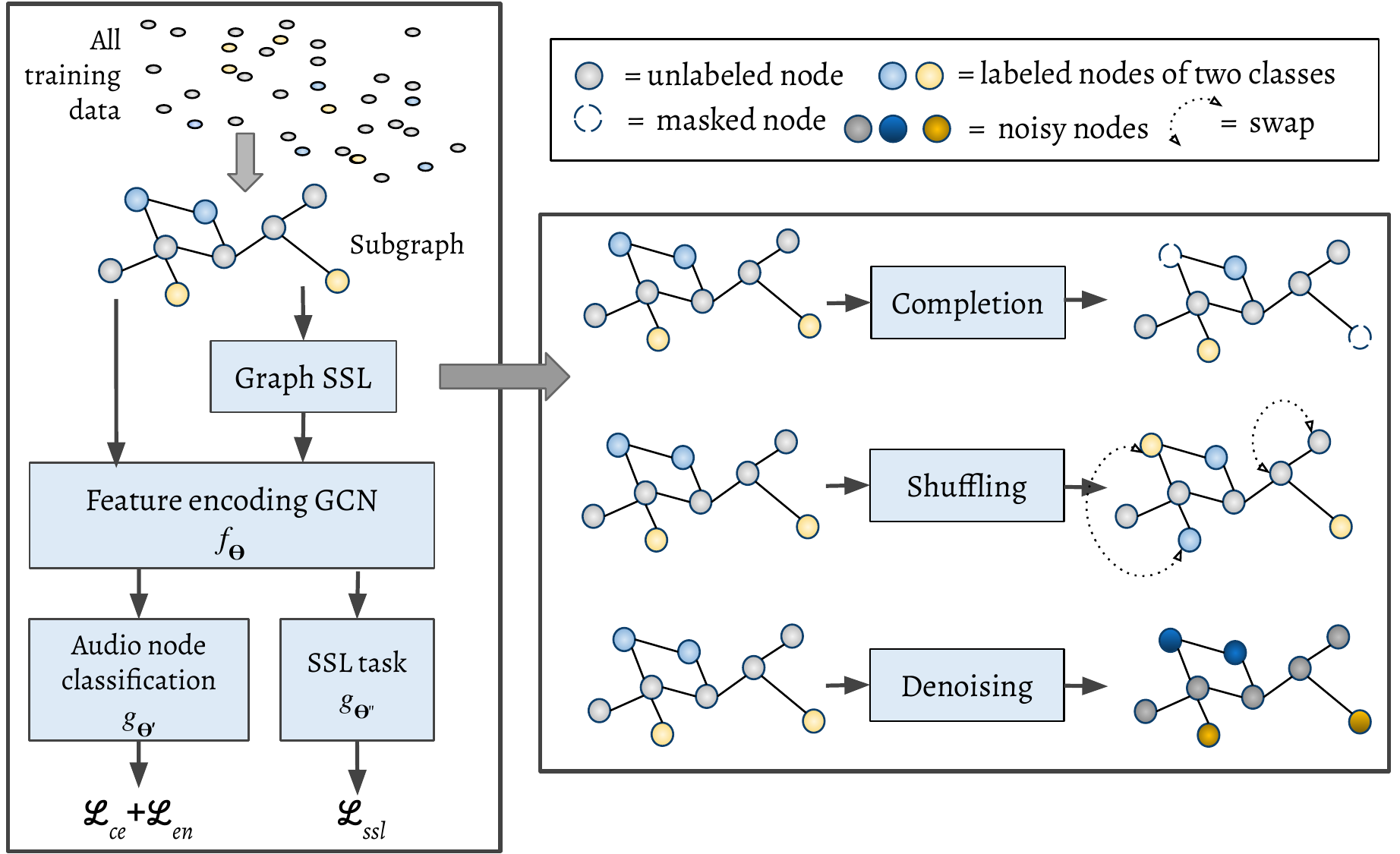}
  \caption{Our model: A subgraph-based audio representation learning framework with SSL task. Our subgraph construction technique is efficient, can handle class-imbalance and the SSL framework facilitates robust and effective learning from highly limited labeled data. }
  \label{fig:main}
\end{figure*}
\par
In this paper, we propose a graph SSL approach to learning effective audio representations from limited amount of labelled data. Considering each audio sample as a graph node, we propose a subgraph-based learning framework with new self-supervision tasks. Our framework takes advantage of the entire pool of available data (labelled and unlabeled) during training; while during inference, our subgraphs are constructed using random edges with no overhead (e.g., nearest neighbour computation) of graph construction. In contrast to the more common SSL-then-finetune approach, we use an auxiliary learning paradigm where an SSL task and a node labelling task are performed jointly. Evaluation on large benchmark databases shows that our model achieves better results than fully supervised models outperforming state-of-the-art on several databases. To summarize, our contributions are as follows: 
\begin{itemize}
    \item We develop a subgraph-based auxiliary learning framework for audio representation learning with limited labelled data. To the best of our knowledge, ours is the first work on self-supervised semi-supervised audio representation learning with graphs.
    \item We propose a new graph SSL task, namely graph shuffling, and a new variant of graph denoising SSL task. We show that they can improve the performance of any graph network for semi-supervised node classification. 
    \item We demonstrate the superior performance of our model for two tasks (acoustic event classification, and speech emotion classification) on three large benchmark audio databases. Our model, despite using limited labelled data, performs better or on par with fully supervised models, and can produce representations that are robust to various types of audio noise.
\end{itemize}

\section{Related Work}
\label{sec:related}
In this section, we review past works in the areas of self-supervised learning and graph neural networks (GNN) in the context of audio processing.
\subsection{Self-supervised learning in audio} 
SSL has gained considerable popularity since its introduction in natural language processing \cite{DevlinCLT19} and computer vision \cite{doersch2015unsupervised,chen2020big,xie2020self} owing to its ability to learn effective data representations without requiring manual labels. 
Acquiring detailed manual labels is arguably more difficult (and often, expensive) in many audio and speech processing tasks, which makes SSL an increasingly popular paradigm in audio analysis. Contrast predictive coding (CPC) was the earliest work on SSL in the audio domain \cite{oord2018representation}. This work demonstrated the applicability of contrastive SSL to audio by predicting latent audio features at a future time instant. The popular model, Wav2Vec, further refined this CPC approach \cite{SchneiderBCA19} to produce state-of-the-art audio presentations.  

The majority of SSL works in audio domain rely on extracting audio descriptors and then utilising deep models to reconstruct a perturbed version of those descriptions as SSL tasks. For example, a self-supervised neural voice synthesizer was used \cite{choi2021neural} to reconstruct the input as an SSL (pretext) task. The Problem Agnostic Speech Encoder (PASE) \cite{PascualRSBB19} is another recent work that seeks to learn multitask speech representations from raw audio by predicting a number of handcrafted attributes like MFCCs and prosody features. Teacher-student models have also been investigated, where the trained model from the previous epoch serves as the teacher for the current epoch \cite{KumarI20}. Several recent papers in audio analysis report that SSL can improve over fully supervised models \cite{tagliasacchi2020pre,hsu2021hubert,niizumi2021byol} while others use crossmodal self-supervision from visual domain \cite{shukla2020does, nagrani2020disentangled}.
\subsection{Graph neural networks in audio}
The predominant approach to audio/speech classification has been using Convolutional Neural Network (CNN) or Long-Short Memory Network (LSTM)-based models on a set of low-level descriptors. Works that use graphs to learn audio representation are limited, but steadily increasing. In a recent work, we have shown that graphs can be used to model audio samples effectively, leading to light-weight yet accurate models for emotion recognition in speech \cite{ShirianG21}. This work used a simple cycle and line graph to describe a given audio data sample. A follow-up work generalised such graph representation of audio to a learnable graph structure \cite{shirian2021dynamic}. A graph-based neural network was utilised to capture the relationships within various speech segments of speakers in a conversation for speech emotion classification \cite{ishiwatari2020relation}. In another work, speech signals are represented as graphs to better capture the global feature representation for speech emotion recognition \cite{liu2021graph}, where deep frame-level features are generated by an LSTM followed by a GNN to classify the graph representation of utterances. In another recent work \cite{tzirakis2021multi}, each audio channel is viewed as a node while constructing a speech graph for speech enhancement task. This allows for the discovery of spatial correlation among several channels. GNNs have been also employed in the context of fusing information from multiple heterogeneous modalities \cite{nie2020c,tang2021fusion}.
\par In a recent study \mbox{\cite{sun2020ontology}}, an ontology-aware approach to acoustic event classification has been proposed that uses feedforward ontology layers and GCNs as two subnetworks. The intra-dependencies among labels are captured by the feedforward ontology layers, while GCN layers focus on modelling the inter-dependency structure of labels. Another work \cite{wang2020modeling} considers the likelihood of co-occurrences between acoustic events by first extracting audio features using a CNN-based network. Then, based on the frequency of audio node labels, a graph structure is created, with each node representing a label. The created graph is then used to train a GCN to learn node representations by propagating information between neighbouring nodes.
Multitask GCN has also been used in literature \cite{shrivastava2020mt} to alleviate the effect of label noise and utilise the hierarchical structure with successful results on audio tagging.
Different from the previous studies, our current paper considers each audio sample as a graph node and presents a subgraph-based self-supervised, semi-supervised learning approach.
\subsection{Graphs and SSL} SSL has been extended to graph data for both fully supervised \cite{tsitsulin2018sgr} and semi-supervised \cite{you2020does,zhu2020self,jin2020self} tasks. Graph SSL tasks usually involve learning the local or global structure, or the context information in the data \cite{jin2020self,tsitsulin2018sgr, you2020does}. Conventional graph tasks such as colouring and partitioning have already been used as SSL tasks \cite{you2020does}. However, we are not aware of any work that uses a graph approach to learning audio representation using SSL.

\section{Proposed Approach}
\label{sec:proposed}
 In this section, we propose our self-supervised (sub)graph-based audio representation learning model. Our model consists of an audio feature encoder, a subgraph construction step and a multitask-SSL architecture with new pretext tasks and loss functions.
\subsection{Audio feature encoder} 
 Our model has a feature encoder $f: \mathcal{S}\rightarrow \mathcal{Z}$ that takes raw audio $\mathcal{S}$ as input and returns embedding $\mathcal{Z}$. These embeddings are used as node attributes in graph $\mathcal{G}$ that we construct using labelled and unlabeled training data (described below). Owing to the different types of audio data (speech and sound), we use two different feature encoders: low-level descriptors for speech data and log-spectrogram based convolutional features for the generic (non-speech) audio. We chose simpler embeddings so as to demonstrate the effectiveness of our graph approach. Nevertheless, our model is not tied to any specific embedding, and rich audio embeddings such as \emph{wav2vec} \cite{SchneiderBCA19} may lead to better classification results. More details about $f$ are provided in the Experiments section.
\subsection{Graph construction during training} 
Given a collection of $N$ (labelled and unlabeled) audio samples for training, we construct an undirected graph $\mathcal{G} = (\mathcal{V},\mathcal{E})$ to capture the relationships among the samples, where $\mathcal{E}$ is the set of all edges between the connected nodes, and $\mathcal{V} = \mathcal{V}_l \cup \mathcal{V}_u$ has $\vert\mathcal{V}_l\vert = M$ labeled and $\vert\mathcal{V}_u\vert = (N-M)$ unlabeled nodes. To construct the graph $\mathcal{G}$, we first consider the labeled nodes. For a node $v_i \in \mathcal{V}_l$, we compute its $k$-nearest neighbors among \emph{all} nodes in $\mathcal{V}_l$ based on their node attributes $\mathbf{z}_i$. We add an edge $e_{ij}\in\mathcal{E}$ with edge weight $a_{ij}= 1$ to the first $Q$ nodes, if $v_j$ is among the $k$-nearest neighbors of $v_i$ and has the same label as $v_i$. We add another edge $e_{ip}\in\mathcal{E}$ with weight $a_{ip}=-1$ between $v_i$ and its farthest node $v_p\in \mathcal{V}_l$. This negative weight is expected to force the two nodes to be apart in the embedding space. Every unlabeled node in $\mathcal{V}_u$ is connected to its two nearest and one farthest neighbors with respective edge weight of $1$ and $-1$, where the neighbors can be any node in $\mathcal{V}$. We thus obtain a corresponding graph adjacency matrix $\mathbf{A} = \{a_{ij}\}_{i,j = 1}^N$. 
\subsubsection{Subgraph construction:} Instead of constructing one large graph containing all training samples, we propose to construct and train on subgraphs. In our case, subgraph construction ensures that we do not end up with a \emph{sparse graph}. This is also motivated by the observations made in several recent papers where subgraphs are able to learn local context more effectively (without oversmoothing or node dependence) while reducing computational load \cite{alsentzer2020subgraph, cong2020minimal, Zeng2020GraphSAINT, huang2020graph}. To construct a subgraph $\mathcal{G}_s$, we randomly select $N_s\in \mathcal{V}_l$ labeled nodes (equal number of samples from each class) and $M_s \in \mathcal{V}_u$ unlabeled training nodes, yielding a set of $\mathcal{V}_s$ nodes, $\vert\mathcal{V}_s\vert = N_s+M_s$. This procedure ensures the degrees of the nodes do not vary too much and class balance is maintained in each subgraph. Next, the edges in the subgraph are added the same way as for the full graph mentioned above. We also show that this approach produces better results than working with a single large graph.
\subsection{Subgraph SSL training with limited labels} 
We adopt an auxiliary learning paradigm to merge self-supervision into the main task of audio classification. This is done by jointly optimizing for node classification and an auxiliary graph SSL task. Our model (see Fig.~\ref{fig:main}) has a shared GCN module for learning the latent audio representations, which is followed by two branches: one for audio classification and the other for SSL. Our model is \emph{inductive}, i.e., neither attributes nor edges of the test nodes are present during the training process. This is a more challenging scenario than transductive learning \cite{HamiltonYL17}.

Our node (audio) classification subnetwork uses supervision from only the true labels while producing \emph{pseudolabels} for the unlabeled nodes. We train this sub-network using two loss functions: \\
(i) Cross-entropy loss $\mathcal{L}_{ce}$ computed for the \emph{labeled} nodes.
\begin{equation}\label{eq:ce}
\begin{aligned}
  \mathcal{L}_{ce} = -\displaystyle\sum_{v_p \in \mathcal{V}_l} \mathbf{y}_p \log(\mathbf{\hat{y}}_p)
\end{aligned}
\end{equation}
(ii) For the \emph{unlabeled} nodes, we propose to compute an \textbf{entropy regularization loss} $\mathcal{L}_{en}$. This can be considered as a measure of class overlap - the lower the entropy loss, the more distinguishable the predicted class labels are.
\begin{equation}\label{eq:e}
\begin{aligned}
  \mathcal{L}_{en} = -\displaystyle\sum_{v_p \in \mathcal{V}_u} P(\mathbf{\hat{y}}_p) \mathrm{log}(P(\mathbf{\hat{y}}_p))
\end{aligned}
\end{equation}
where, $\mathbf{y}_p$ is the true label for node $v_p$, $\mathbf{\hat{y}}_p$ is its predicted label, $\mathcal{V}_l$ and $\mathcal{V}_u$ are the sets of labeled and unlabeled training samples. The SSL subnetwork is trained using a graph SSL task optimizing over a loss function $\mathcal{L}_{ssl}$ (discussed below). Given a subgraph input $\mathcal{G}_s$, the overall optimization is given by: %
\begin{equation}\label{eq:overall_loss}
\begin{aligned}
    \underset{\Theta,\Theta', \Theta''}{\text{min }}{\big [\mathcal{L}_{ce}(\Theta,\Theta', \mathcal{G}_s) + \lambda_1\mathcal{L}_{en} (\Theta,\Theta', \mathcal{G}_s)+} \\ {\lambda_2\mathcal{L}_{ssl}(\Theta,\Theta'', \hat{\mathcal{G}}_s)\big ]}
\end{aligned}
\end{equation}
where $\Theta$, $\Theta'$, and $\Theta''$ are the learnable parameters for the shared GCN, classification GCN and the SSL sub-networks, $\hat{\mathcal{G}}_s$ is the SSL variant of $\mathcal{G}_s$, and $\lambda_1$ and $\lambda_2$ control the relative weights of SSL loss and entropy regularization. 
\begin{algorithm}[tb]
\SetAlgoLined
\SetKwInOut{Output}{Output}
\SetKwInput{Input}{Input~}
\SetKwProg{Fn}{Function}{}{end}
\Input{Labeled nodes $\mathcal{V}_l$, unlabeled nodes $\mathcal{V}_u$, node embeddings $\mathbf{Z}$} 
\Output{Learned parameters $(\Theta,\Theta',\Theta'')$, pseudolabels $\hat{\mathbf{y}}_p$ for $\mathcal{V}_u$ }
\BlankLine

\For{each epoch} 
{
$\mathcal{G}_s \leftarrow \texttt{subgraphConstruct}(\mathcal{V}_l,\mathcal{V}_u,\mathbf{Z})$\\
$\hat{\mathcal{G}}_s \leftarrow \texttt{createGraphforSSL}(\mathcal{G}_s)$\\

$\hat{\mathbf{y}} \leftarrow g_{\Theta^{'}}(f_\Theta(\mathcal{G}_s))$; \, \,\,
$\tilde{\mathbf{Z}} \leftarrow  g_{\Theta^{''}}(f_\Theta(\hat{\mathcal{G}}_s))$\\

\vspace{1mm}
$\Theta,\Theta', \Theta'' \leftarrow \mathcal{L}_{ce} + \lambda_1\mathcal{L}_{en} + \lambda_2\mathcal{L}_{ssl}$\\
} 

\Fn{\emph{\texttt{subgraphConstruct}} $(\mathcal{V}_l,\mathcal{V}_u,\mathbf{Z}$)}{
$\mathcal{V}_s$ $\leftarrow$ randomly select $N_s$ nodes from $\mathcal{V}_l$ and $M_s$ nodes from $\mathcal{V}_u$\\
 \For{$\forall v_s \in \mathcal{V}_s$}
 {
  $\mathcal{V}_{nn} \leftarrow$ \text{nearest neighbors}$(v_s)$
  
  \eIf{$v_s \in \mathcal{V}_l$}{
    $\mathcal{E}_s\overset{+}{\leftarrow}$ edge between $v_s$ and $2$ nearest nodes $v_{i}\in\mathcal{V}_{nn}$ with same labels with edge weight $1$ }

       { $\mathcal{E}_s\overset{+}{\leftarrow}$ edge between $v_s$ and $2$ nearest nodes $v_{i}\in\mathcal{V}_{nn}$ with edge weight $1$}

   }
    $\mathcal{E}_s\overset{+}{\leftarrow}$ edge between $v_s$ and farthest $v_{i}\in\mathcal{V}_{nn}$ with weight $(-1)$
 } \textbf{return} $\mathcal{G}_s(\mathcal{V}_s,\mathcal{E}_s)$
 \caption{Subgraph-based SSL training}
\end{algorithm}

\par To this end, we experiment with three graph SSL tasks. These SSL tasks are \emph{model-agnostic} and can be used with any graph neural network. We propose a novel proxy task in the graph SSL domain: graph shuffling. In addition, we also experiment with the recently introduced graph completion and graph denoising proxy task. 

\subsubsection{Graph denoising} Motivated by past works \cite{lu2013speech,fatemi2021slaps}, we employ a new variant of the SSL task of graph denoising. Given a subgraph $\mathcal{G}_s$, we construct a noisy graph $\Hat{\mathcal{G}}_s$, where Gaussian noise is added to every node feature vector $ \hat{\mathbf{z}}_s(i)= \mathbf{z}_s (i) + \mathbf{x}(i)$, where $\mathbf{x}(i)\sim\mathcal{N}(0,\epsilon)$ 
by adding a Gaussian noise with zero mean and $\epsilon$ variance to each node feature. The SSL regression task is to learn to reconstruct node feature matrix $\mathbf{Z}_s$ from noisy $\hat{\mathbf{Z}}_s$ by optimizing the following loss function:
\begin{equation}\label{eq:L_graph_deno}
    \mathcal{L}_{ssl} = \frac{1}{|\mathcal{V}_{s}|} \Vert \Tilde{\mathbf{Z}}_s - \mathbf{Z}_s \Vert_F^2 \text{\, where\,\,} \tilde{\mathbf{Z}}_s = g(\Theta, \Theta'', \hat{\mathbf{Z}}_s)
\end{equation}

It is worth mentioning that in a recent study \mbox{\cite{fatemi2021slaps}}, a denoising autoencoder has been used to perform SSL on graph data. This work is different from ours; they used a self-supervised approach with cross-entropy loss that jointly learns the graph structure and node features at the same time. while we apply the cleaning loss (Eq. \mbox{\ref{eq:L_graph_deno}}) in a fixed graph after a graph-based feature extractor to reconstruct the node features.

\subsubsection{Graph completion:}
This SSL task forces the network to learn to reconstruct missing information so as to learn local context. Following a recent work \cite{you2020does}, we mask a random set of target nodes $\mathcal{V}_{sc}\subset\mathcal{V}_s$ by setting their node attributes to zero. The task is to recover this missing information for the target nodes. Given $\mathcal{G}_s$, denote the ground truth feature matrix corresponding to $\mathcal{V}_{sc}$ as $\mathbf{Z}_{sc}$ and its predicted version as $\Tilde{\mathbf{Z}}_{sc}$, the SSL loss is then given by
\begin{equation}\label{eq:L_graph_comp}
    \mathcal{L}_{ssl} = \frac{1}{|\mathcal{V}_{sc}|} \Vert \Tilde{\mathbf{Z}}_{sc} - \mathbf{Z}_{sc} \Vert_F^2
\end{equation}
\subsubsection{Graph shuffling (proposed):} 
We propose a novel SSL task that aims to determine whether or not a graph node is in its correct position. This task encourages the graph network to learn structural dependencies among nodes without using the available labels. Given $\mathcal{G}_s$, we randomly sample a set of graph nodes $\mathcal{V}_{sh}\subset\mathcal{V}_s$ and shuffle their node attributes randomly with each other creating $\hat{\mathcal{G}}_{s}$. The SSL task is posed a binary node classification task on $\hat{\mathcal{G}}_{s}$ where the model outputs 1 if is node is unchanged and 0 otherwise.
\begin{equation}\label{eq:L_graph_shuffle}
    \mathcal{L}_{ssl} =-\frac{1}{\vert \mathcal{V}_{sh}\vert} \displaystyle\sum_{v_{i} \in \mathcal{V}_{sh}} {y}_{i} \log({\hat{y}}_{i}) + (1-{y}_{i}) \log(1-{\hat{y}}_{i})
\end{equation}
\subsection{Subgraph construction during inference} 
After completing training, we obtain a large number of \emph{pseudolabeled} samples. In order to create the subgraph during inference, we randomly sample $\vert \mathcal{V}_s \vert$ nodes (equal number of nodes from each class) from the entire set of training samples considering both true and psesuolabels. The edges between these nodes are constructed as was done during training. Each test node (audio sample) $v_t$ has to be connected to this graph structure $\mathcal{G}_s$. Computing nearest neighbours during inference time may not always be practical; hence, we propose to connect $v_t$ with $T$ training nodes (labelled or pseudolabeled) \emph{randomly} with edge weight $1$.

Please note that during interference time, we do not need to compute nearest neighbors of the test nodes as the test nodes are connected randomly to the nodes from the training set. For the nodes in the inference subgraph coming from the training set, the nearest neighbours only come from the training set - information we already have precomputed and stored from the beginning. In summary, no new nearest neighbour computing is needed during inference. This is by design, precisely to avoid storing embeddings/labels and computing nearest neighbours again. Therefore we only need to store the pairwise distances of the training nodes in the RAM, not embeddings or labels. This makes our method lightweight, which may be suitable even for edge devices.

\subsubsection{Optimal number of random edges} A natural question is what is the optimal value of $T$ so as to ensure that we do not connect to only pseudolabeled nodes, which could be incorrect. Hence, we ask: \emph{What is the minimum number of nodes a test node, $v_t$, should be connected with, such that there is at least one connection with a true-labelled node?} 

Let $\mathcal{G}$ be a graph with $\vert \mathcal{V}\vert$ nodes including known $N$ true labelled and $M$ pseudolabeled nodes. The probability of having all $T$ edges from $v_t$ to be connected with only the pseudolabeled nodes is given by $\frac{\binom{M-1}{T}}{\binom{N+M-1}{T}}$ using Hypergeometric distribution. Therefore, the probability of an $v_t$ to be connected to at least one true labeled node is given by $P = 1- \frac{\binom{M-1}{T}}{\binom{N+M-1}{T}}$. With known $N$ and $M$, we set a high value of $P (= 0.9)$ to compute the value of $T$ for our experiments. Once the graph is thus constructed, we use only the audio classification branch to determine the class labels for the test nodes.

%
\section{Experiments}
\label{sec:Exp}
This section presents extensive experimental results on the analysis of our model and demonstrates its effectiveness for audio classification tasks.
\subsection{Semi-supervised acoustic event classification}
\subsubsection{Datasets} 
\begin{figure}[tb]
  \centering
    \fbox{\includegraphics[width=0.8\linewidth, trim=0mm 0cm 0mm 0mm, clip=true, bb=0 0 236 272]{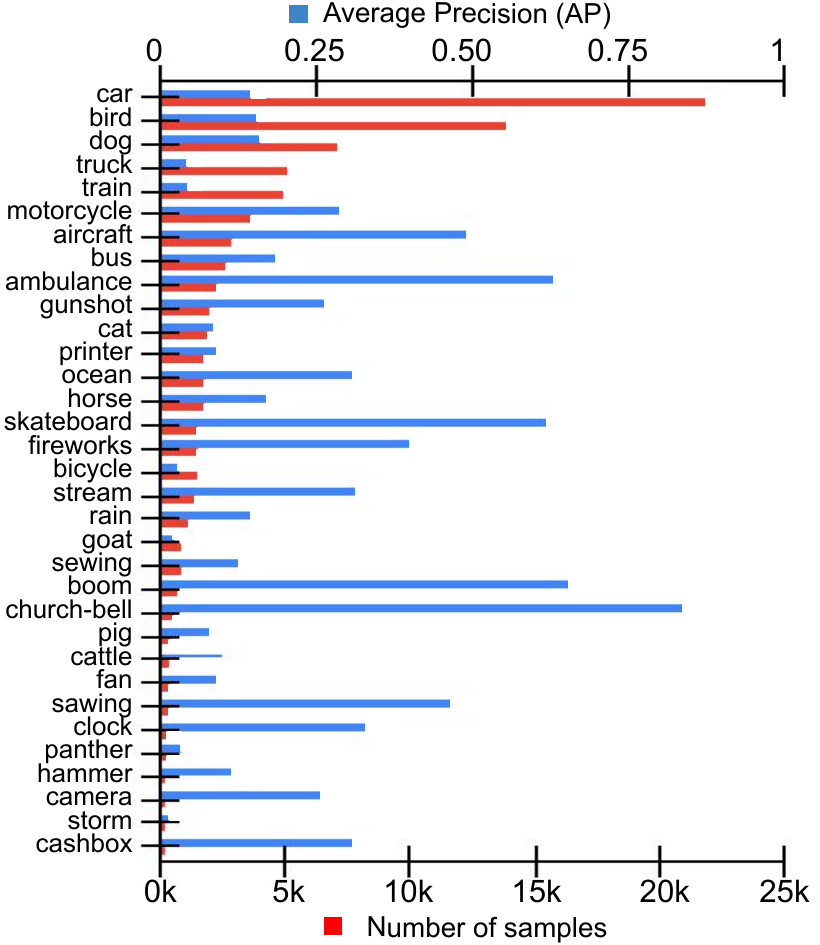}}
    \caption{Results on \textbf{AudioSet}: Class distribution and average precision per class achieved using our model with graph completion task. Note that our model can achieve high AP even for classes with fewer samples owing to our graph construction process.}\label{fig:audioset_AP}
    \vspace{-4mm}
\end{figure}

We use a large scale weakly labeled database called the \textbf{AudioSet} \cite{gemmeke2017audio} that contains audio segments from YouTube videos. We work with 33 class labels that have a high rater confidence score ($\geq 0.7$) (see Fig.~\ref{fig:audioset_AP} for the names of those classes). This yields a training set of around 89,000 audio clips and a test set of more than 8,000 audio clips. We consider only 10\% of the 89,000 training samples as labelled and the rest are used as unlabeled training data for our experiments.
\subsubsection{Feature encoder} To extract the node features, each audio clip is divided into non-overlapping 960 ms segments. For each segment, a log-mel spectrogram is computed by taking its short-time Fourier transform using a frame of length 25 ms with 10 ms overlap, 64 mel-spaced frequency bins and log-transforming the magnitude of each bin. This creates log-mel spectrograms of dimension $96\times64$ which are the input to the pre-trained VGGish network \cite{hershey2017cnn}. 
We use the 128-dimensional features extracted from the VGGish for each log-mel spectrogram and average over all segments to form the final vector representation of each audio clip. Note that although we use VGGish embeddings to be comparable with previous works, other generic audio embeddings will work as well.
\subsubsection{Experimental settings}
To construct the subgraph during training, we sample data with $M_s = 5$, $N_s = 2\times \text{number of classes}$, and $T=4$. It is worth noting that the $T$ value is calculated using the probabilistic equation in section III.D, with $P>0.99$. We also tested this hypothesis experimentally. From $T=1$ to $20$, we steadily increase the value of $T$. Each step improves classification performance by a significant margin until $T=4$, where we get $63.8\%$ accuracy. Increasing the $T$ value further does not result in any significant performance improvement. This yields a subgraph size of 71 nodes. This process ensures the subgraph nodes are class-balanced every time. The subgraph construction process is repeated until all unlabeled nodes appear at least once i.e., the subgraphs are constructed by sampling without replacement. Note that for constructing the edges, the nearest neighbours need to be computed only once for all the training nodes. For the test nodes, no nearest neighbour computation is needed due to our random edge construction strategy.
For graph neural network, we select regular GCN \cite{kipf2017semi} with $2$ graph convolution layers and a hidden size of $256$ for all layers. We used the same architecture for all experiments.
We use the 80:10:10 train:validation:test split \emph{only} for the semi-supervised framework, where we consider $10\%$ of that $80\%$ train data as labelled and the rest as unlabeled. All hyperparameters were chosen solely based on validation data, with test data accounting for no model or parameter selections.
In order to obtain robust estimates of performance metrics, this process is repeated 5 times to report the average accuracy and standard deviation. We set $\lambda_1=0.01$ and $\lambda_2=0.1$ (see Eq. \ref{eq:overall_loss}). Our model uses Xavier initialization and the Adam optimizer with a learning rate of $0.001$ for all experiments. We use Pytorch for implementing our models and the baselines. All models were trained on a single NVIDIA RTX-2080Ti GPU.
\subsubsection{Baselines} We compare our method with a number of fully supervised models in Table \ref{tab:Auco_Eve_Clas}. The Spectrogram-VGG model is the same as configuration A in \cite{simonyan2014very} with only one change: the final layer being a softmax with 33 units. The feature for each audio input to the VGG model is a log-mel spectrogram of dimension 96$\times$64 computed averaging across non-overlapping segments of length 960ms. 
We did not adjust the hyperparameters for baselines other than VGG and used the settings stated in the original papers.
The fully supervised version of our model follows the same graph construction strategy as proposed with 80:10:10 (train:validation:test) split. The split process is done 5 times for our models and average performance with standard deviation is reported. All other baseline implementations are done by the authors and follow the same evaluation protocol as the proposed methods. 
\begin{table}[t]
\centering
\caption{Acoustic event classification results on \textbf{AudioSet}: Our model, using only 10\% labeled data, outperforms all semi-supervised models and several fully supervised models. The fully supervised version of our model without SSL shows comparable performance to the state of the art, indicating the effectiveness our subgraph-based learning strategy for audio classification.Please note that the parameters for our model excludes the feature extractor, which varies depending on the features used.}
\label{tab:Auco_Eve_Clas}
\renewcommand*{\arraystretch}{1.0}
  \begin{tabular}{l|c|c}
        \toprule
        \bf Model & \bf mAP & \bf Params          \\ \midrule
        \multicolumn{3}{c}{\emph{Semi-supervised}}\\
        \midrule
        Ours w/o SSL                 & $0.23\pm0.01$ & $218$K\\   
        Ours w/ denoise            & $0.26 \pm0.00$ & $260$K\\
        Ours w/ completion          & \bf $0.27\pm 0.01$ & $260$K\\
        Ours w/ shuffle           & $0.24\pm0.00$ & $219$K\\
        Ours w/ all three SSL    & $\mathbf{0.28}\pm0.02$ & $261$K\\
        \midrule
        Spectrogram-VGG   &$0.16\pm0.05$  & $6$M \\ 
        AST \cite{gong2021ast} & $0.22\pm0.01$   & $88$M   \\ 
        \midrule
        \multicolumn{3}{c}{\emph{Fully supervised}}\\
        \midrule
        Ours w/o SSL   & $0.42\pm0.02$  & $218$K  \\
        Spectrogram-VGG   & $0.26\pm0.01$ & $6$M\\ 
        DaiNet \cite{dai2017very} & $0.25\pm0.07$ & $1.8$M \\
        Wave-Logmel \cite{kong2020panns} & $0.43\pm0.04$  & $81$M  \\
        VATT \cite{akbari2021vatt} & $0.39\pm0.02$  & $87$M  \\
       AST \cite{gong2021ast} & $\mathbf{0.44}\pm0.00$  & $88$M  \\
        \bottomrule
        \end{tabular}
        \vspace{-4mm}
\end{table}
\begin{table}[t]
    \caption{Speech emotion recognition results: Classification (unweighted) accuracy (in \%) on two benchmark databases are presented. Our model, using only $10\%$ labeled data, outperforms semi-supervised and several fully supervised models on both databases. The fully supervised version of our model produces the highest accuracy even without SSL, indicating the effectiveness of our subgraph-based learning strategy. (* indicates audiovisual models).}
    \label{tab:SER_res}
    \centering
    \renewcommand*{\arraystretch}{1.0}
    \resizebox{1.0\linewidth}{!}{
      \begin{tabular}{l|c|c|c}
        \toprule
       \bf  Model   & \small{IEMOCAP}   & \small{MSP-IMPROV} & \bf  Param          \\ \midrule
    \multicolumn{4}{c}{\emph{Semi-supervised}} \\ \midrule
        Ours w/o SSL    & $63.8\pm2.2$  & $58.6\pm1.8$  & $212$K \\
        Ours w/ denoise & $68.0\pm1.1$  & $64.1\pm1.0$  & $271$K\\ 
        Ours w/ completion  & $66.4\pm1.7$  & $63.8\pm1.5$  & $271$K\\ 
        Ours w/ shuffle & $65.9\pm1.4$  & $64.1\pm1.3$  & $213$K\\
        Ours w/ all three SSL   & $\mathbf{68.6}\pm1.2$ & $\mathbf{65.2}\pm1.8$   & $272$K\\
        \midrule
        LadderNet \cite{huang2018speech}    & $60.7$    & - & -\\
        Transformer* \cite{liang2020semi}   & $61.2$    & - & -\\
        SimCLR \cite{jiang2020speech}   & $65.1$    & - & $30$M          \\
        \midrule
        \multicolumn{4}{c}{\emph{Self-supervised (non-graph)}} \\ \midrule
        SSAST \cite{gong2021ssast}  & $59.6$    & - & $89$M\\
        BYOL-S/CvT \cite{scheidwasser2021serab} & $65.1$    & - & $5$M\\
        Wav2vec2.0 \cite{yang2021superb} & $65.6$  & - & $317$M\\
        HuBERT \cite{yang2021superb} & $67.6$   & - & $316$M\\
        \midrule
        \multicolumn{4}{c}{\emph{Fully supervised}} \\ \midrule
        Ours w/o SSL    & $\mathbf{70.5}$   & $\mathbf{66.7}$   & $212$K \\ 
        SegCNN \cite{mao2019deep}   & $64.5$    & - & -\\
        GA-GRU \cite{su2020improving}   & $63.8$    & $55.4$    & -\\
        
        CNNattn \cite{MaoCKL20} & $66.7$    & - & -\\
        WADAN \cite{yi2020improving}    & $64.5$    & - & -\\
        SpeechGCN \cite{shirian2021compact}  & $62.3$   & $57.8$    & $30$K\\
        \bottomrule
        \end{tabular}
        }
        \vspace{-4mm}
\end{table}
\begin{figure}
    \centering
\begin{minipage}[t]{0.4\textwidth}
\centering
\includegraphics[width=\linewidth, trim=0mm 0mm 0mm 0mm, clip=true]{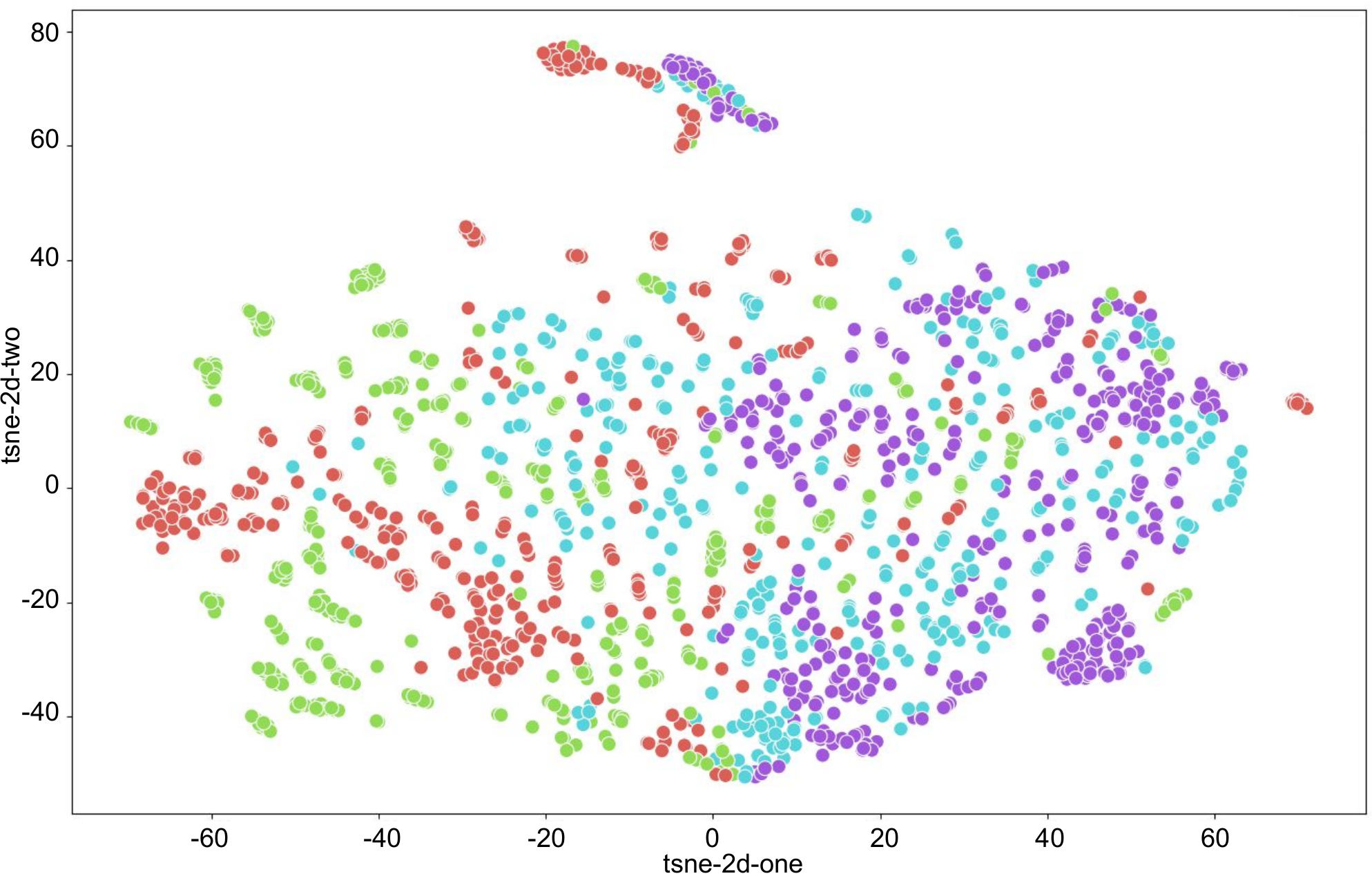}

(a)
\end{minipage}
\hfill
\begin{minipage}[t]{0.4\textwidth}
\centering
\includegraphics[width=\linewidth, trim=0mm 0mm 0mm 0mm, clip=true]{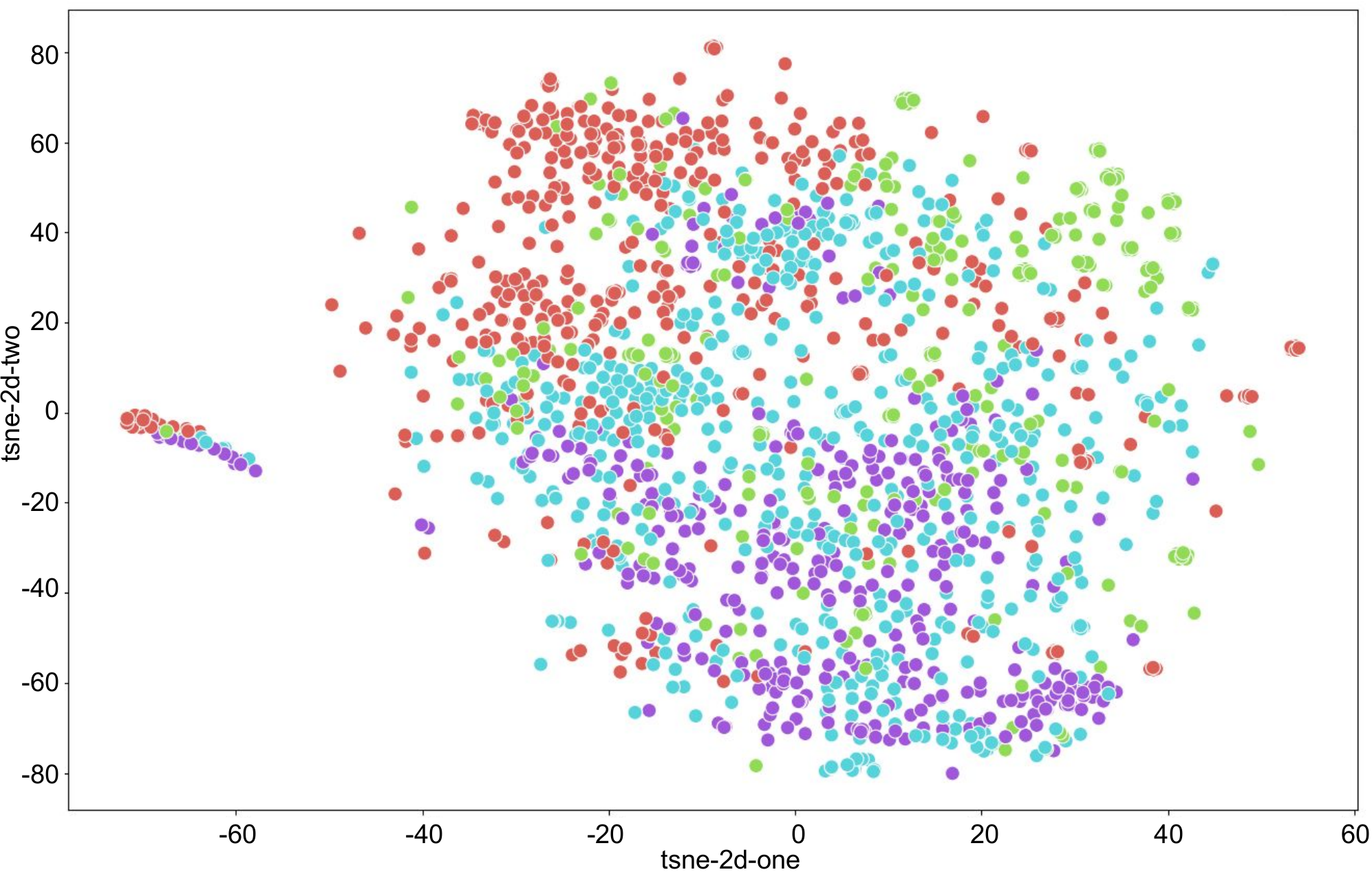}

(b)
\end{minipage}
  \caption{Comparison of t-SNE plots and the Silhouette score for (a) our model (Silhouette score = 0.147) and (b) non-graph model SSAST (Silouette score - 0.013). The Siloutte scores being a measure of clustering quality indicates that our model learns more discriminative embeddings.}
\label{fig:t-sne}
\vspace{-6mm}
\end{figure}
\subsubsection{Results} 
Table \ref{tab:Auco_Eve_Clas} reports the average recognition accuracy (averaged over 5 runs) with standard deviation for each model and their variants. 
For the case of all three SSL, we applied all three self-supervised tasks in parallel with a shared feature encoder and add the corresponding loss functions to the main classification loss.
It compares the performance of our model with different SSL tasks with that of fully supervised models in terms of mean Average Precision (mAP). The graph models with SSL outperform the plain graph model without SSL. When compared with the fully supervised models, our graph SSL models (denoise and completion in particular) outperform Spectrogram-VGG and DaliNet \cite{dai2017very}. Our model also has significantly fewer learnable parameters. The fully supervised GCN model uses the same graph construction method as proposed and performs very close to the state-of-the-art. This demonstrates the effectiveness of our graph construction strategy.
Fig. \ref{fig:audioset_AP} shows the average precision (AP) per class for the AudioSet database. A high AP is achieved even for classes with fewer samples, e.g., \emph{church-bell} has an AP of 0.843 even with only 627 samples. This suggests that our model is not highly affected by a lower number of samples owing to how we construct the sub-graphs during the training process.
\subsection{Semi-supervised speech emotion recognition}
\subsubsection{Datasets} For this task, we use the two most popular speech emotion datasets. \\
The \textbf{IEMOCAP} \cite{busso2008iemocap} dataset contains 12 hours of speech collected over 5 dyadic sessions with 10 subjects. It includes 4490 utterances with labels, 1103 \emph{anger}, 595 \emph{joy}, 1708 \emph{neutral} and 1084 \emph{sad}. \\
The \textbf{MSP-IMPROV} \cite{busso2016msp} contains 7798  utterances from 12 speakers collected across six sessions including 792 samples for \emph{anger}, 3477 for \emph{neutral}, 885 for \emph{sad} and 2644 samples for \emph{joy}. The train-test split is the same.
\subsubsection{Feature encoder} Following relevant past work on speech emotion analysis \cite{schuller2009interspeech}, we extract a set of low-level descriptors (LLDs) from the speech utterances using the OpenSMILE library \cite{eyben2013recent}. This feature set includes Mel-Frequency Cepstral Coefficients (MFCCs), zero-crossing rate, voice probability, fundamental frequency (F0) and frame energy. For each audio sample, we use a sliding window of length 25ms (with a stride length of 10ms) to extract the LLDs. The local features are smoothed temporally using a moving average filter, and the smoothed version is used to compute their respective first-order delta coefficients. Then we compute the mean, max, standard deviation, skew and kurtosis of the extracted LLDs and their delta coefficients to compute one feature vector per speech sample. Altogether this yields node embeddings of dimension 165 for an audio clip.
\subsubsection{Experimental settings}
We follow the same settings as in the acoustic event classification task with a change in the graph size. The input graph used for this task has $53$ nodes, where $M_s=5$, $N_s = (12 \times\text{number of classes})$, and $T=4$. Further discussion on the effect of graph size is presented in the next section.
\subsubsection{Results} Table \ref{tab:SER_res} compares our model against the state-of-the-art supervised, (non-graph) semi-supervised, and (non-graph) self-supervised methods on the two speech databases. Clearly, our method (with denoise SSL in particular) outperforms others by a significant margin, using only $10\%$ of the training data as labelled. Again note that the fully-supervised version of the model yields comparable results with the state-of-the-art.
Furthermore, as compared to non-graph self-supervised methods, our method provides comparable or higher results (especially with denoise and all three SSL) with a substantially reduced number of parameters as shown in Table II.
We also showed graph-based and non-graph (SSAST) features \mbox{\cite{gong2021ssast}} using the t-SNE plot . For comparison, we used the Silhouette score, which is a tool for evaluating clustering quality. As seen in Fig. \mbox{\ref{fig:t-sne}}, our method produces significantly better clusters with a $0.147$ Silhouette score  when compared to the non-graph method with a $0.013$ Silhouette score.
The results show that SSL with subgraph is an effective learning framework for speech classification even when labelled data is highly limited.
\subsection{Model analysis and ablation studies}
\label{subsec:model_analysis} 
\subsubsection{Ablating graph construction}
We next attempt to understand the effectiveness of self-supervision on transductive semi-supervised node classification. This requires experimenting with databases where the graph structures are known so as to disentangle the effect of our proposed graph construction methodology. We use three graph citation databases \textbf{Cora} (2708 nodes, 7 classes) and \textbf{Citeseer} (3327 nodes, 6 classes)  \cite{sen2008collective} and \textbf{Pubmed} (19717 nodes, 3 classes) \cite{namata2012query} 
following the standard benchmarking framework \cite{dwivedi2020benchmarking}.
\par To verify the universality of our SSL tasks, we conduct experiments on several state-of-the-art graph neural networks: (i) Standard GCN \cite{kipf2017semi}, (ii) Graph Attention Network (GAT) \cite{velivckovic2017graph} - a powerful variant of GCN, (iii) Graph Isomorphism Network (GIN) \cite{xu2018powerful} and (iv) GraphMix \cite{verma2019graphmix}. Based on the graph models, we use a joint learning framework where the SSL task as an auxiliary task and node classification is the primary task. Table \ref{tab:bench_graph} compares the performance of the above graph models when augmented with the three SSL tasks. Clearly, the SSL tasks (particularly, denoising and completion) improve the accuracy of all models across all databases cases. As mentioned before, the SSL tasks we consider are model-agnostic and thus can be used to enhance any graph model. Also, note that these experiments use 16-dimensional embedding while the emotion recognition task used 165-dimensional embeddings and acoustic even classification used 1024-dimensional embeddings. This shows that our model works well with different types and dimensions of embeddings.
\begin{table}[t]
\begin{center}
\caption{To show the effectiveness of our proposed framework when the graph structures are \emph{known}, we present semi-supervised node classification results (in \% accuracy) on benchmark graph databases. This essentially disentangles the effect of our proposed graph construction methodology from the SSL-based semi supervised model. The results show that SSL tasks improve over basic models in almost all cases irrespective of the graph network used.} 
\label{tab:bench_graph}
\small
\renewcommand*{\arraystretch}{1.1}
\begin{tabular}{l|c| ccc}
\toprule
  & \bf SSL task & \bf Cora & \bf Citeseer 
  & \bf Pubmed 
  \\
\midrule
\multirow{ 4}{*}{\rotatebox{90}{GCN}}& \xmark & $80.9\pm0.6$         & $70.7\pm0.6$            & ${79.1}\pm0.5$
\\
& denoise & $81.1\pm0.8$         & $71.1\pm0.8$            & $78.4\pm0.8$    
\\
& completion & $\mathbf{81.6}\pm0.8$ &  $\mathbf{71.6}\pm0.6$  & $\mathbf{79.2}\pm0.7$  
\\
& shuffle & $\mathbf{81.6}\pm0.6$   & $70.1\pm1.1$       & $78.4\pm0.6$     
\\
\midrule
\multirow{ 4}{*}{\rotatebox{90}{GAT}}& \xmark  &$83.1\pm0.5$  & $72.1\pm0.6$    &$77.5\pm0.4$ 
\\
& denoise  & $\mathbf{84.2}\pm1.0$  & $\mathbf{73.1}\pm0.5$  &$\mathbf{78.2}\pm0.5$ 
\\
& completion  &$84.2\pm0.4$  & $72.8\pm1.1$  &$78.0\pm0.5$ 
\\
& shuffle &$83.2\pm0.9$  & $72.6\pm0.7$    &$77.7\pm0.4$ 
\\
\midrule
\multirow{ 4}{*}{\rotatebox{90}{GIN}}& \xmark   &$77.2\pm0.5$    & $68.1\pm0.7$   & $77.0\pm0.4$ 
\\
& denoise    &$\mathbf{78.9}\pm0.8$          & $69.1\pm1.4$             & $77.2\pm0.3$ 
\\
& completion    &$78.8\pm0.6$     & $\mathbf{69.8}\pm1.2$   & $\mathbf{77.7}\pm0.3$ 
\\
& shuffle    &$78.6\pm1.1$          & $69.3\pm0.8$          & $77.3\pm0.4$  
\\
\midrule
\multirow{ 4}{*}{\rotatebox{90}{GraphMix}}& \xmark  & $83.8\pm0.8$  &$74.3\pm0.7$    & $80.6\pm0.6$
\\ 
& denoise  & $\mathbf{84.4}\pm0.7$  & $\mathbf{75.2}\pm0.5$   & $81.4\pm0.4$  
\\ 
& completion  & $84.4\pm0.7$  &$74.3\pm0.7$    & $81.2\pm0.3$
\\ 
& shuffle  & $84.2\pm0.4$  &$74.4\pm0.6$    & $\mathbf{81.9}\pm0.3$
\\ 
\bottomrule
\end{tabular}
\end{center}
\vspace{-4mm}
\end{table}
\subsubsection{Why subgraph instead of a large graph} 
\begin{figure}
    \centering
\begin{minipage}[t]{0.4\textwidth}
\centering
\fbox{\includegraphics[width=\linewidth, trim=0mm 0mm 0mm 0mm, clip=true, bb=0 0 535 350]{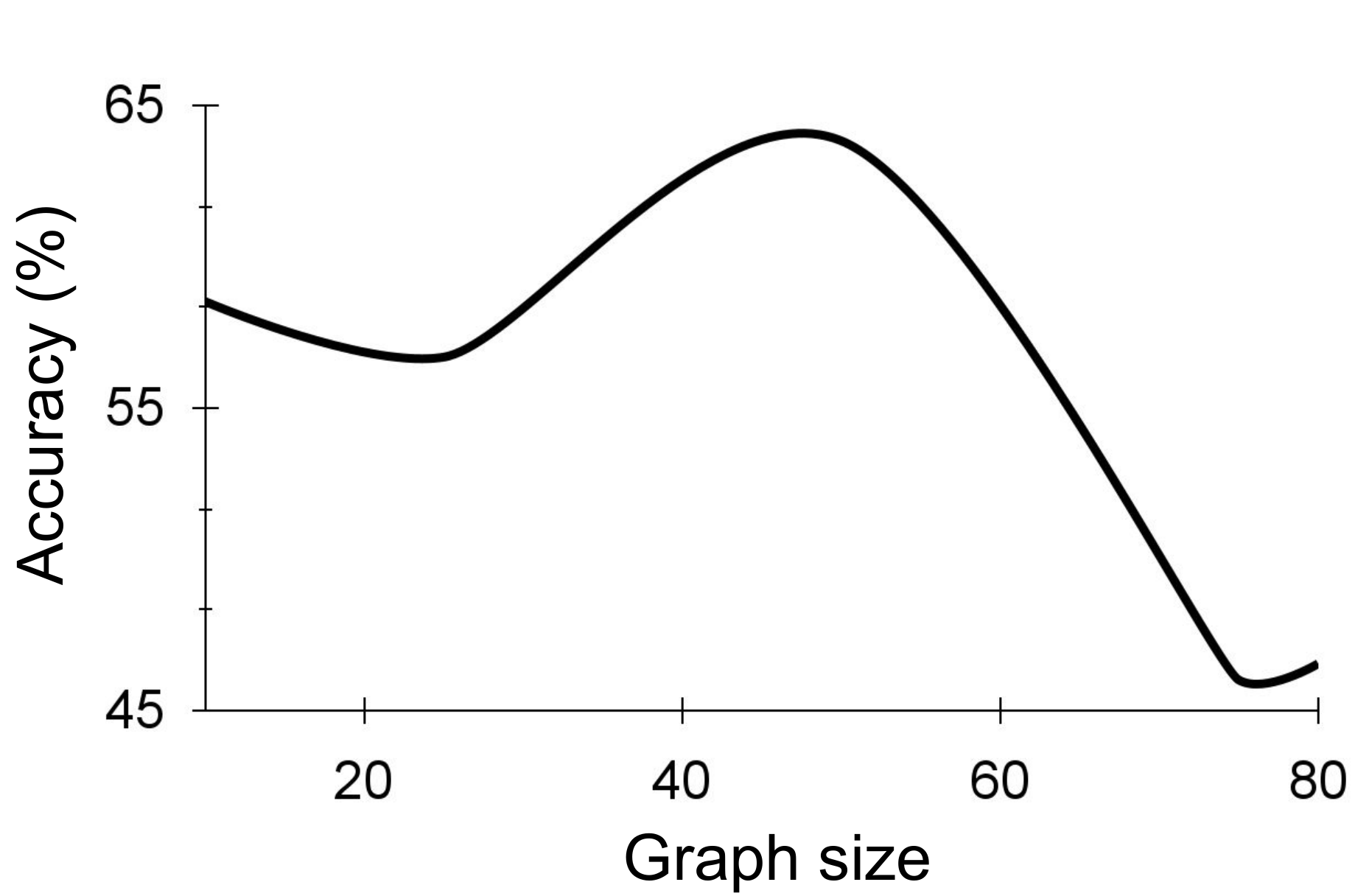}}
\label{fig:graph_size}
(a)
\end{minipage}
\hfill
\begin{minipage}[t]{0.4\textwidth}
\centering
\fbox{\includegraphics[width=\linewidth, trim=0mm 0mm 0mm 0mm, clip=true, bb=0 0 657 421]{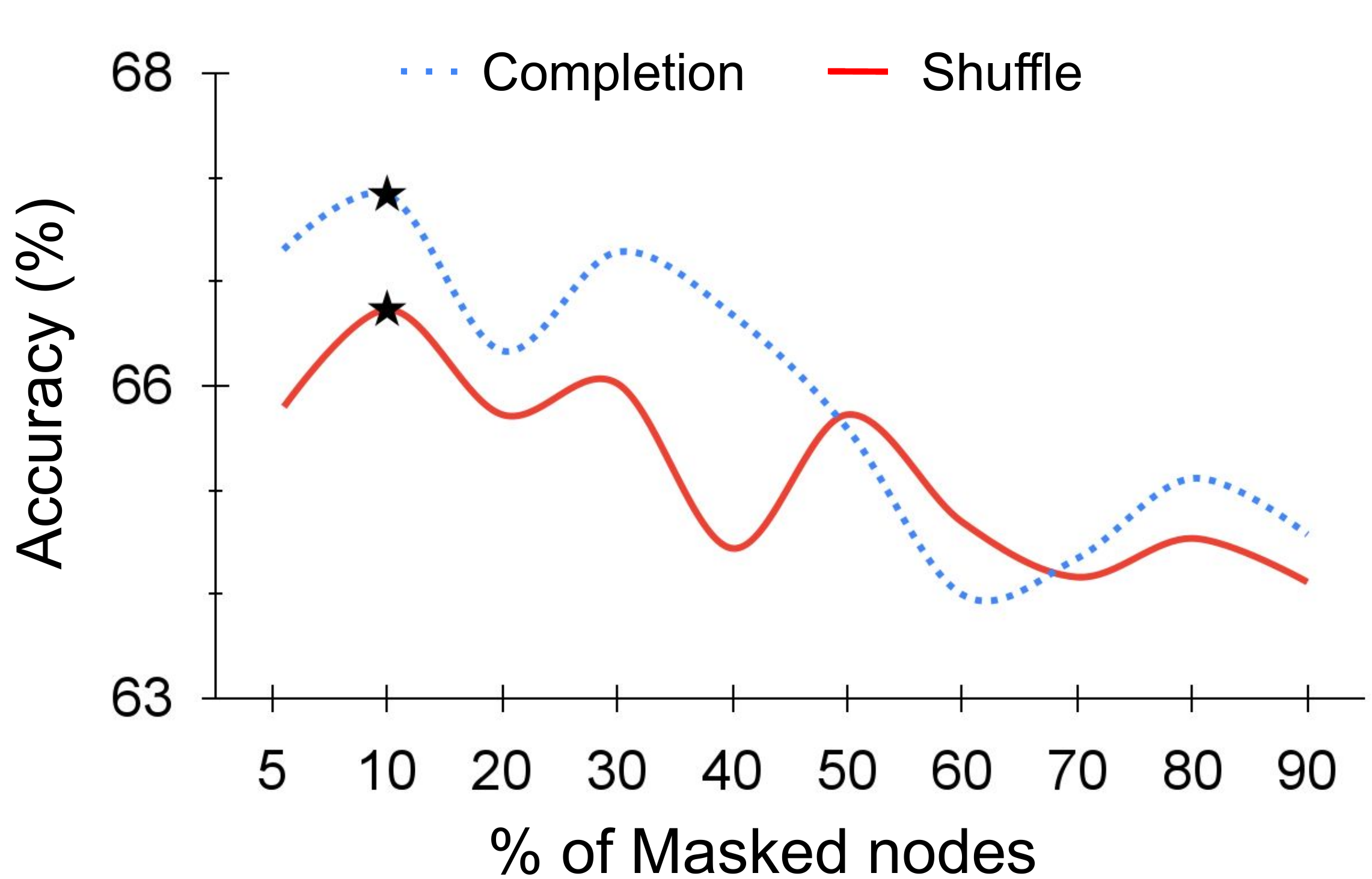}}
\label{fig:mask_ratio}

(b)
\end{minipage}
  \caption{(a) Impact of graph size on audio classification performance on the IEMOCAP database.  (b) Effect of the fraction of masked nodes for the SSL tasks. Results shown are for IEMOCAP database. The $\star$ indicates best performance.}
\label{fig:Hyper_param_analysis}
\vspace{-4mm}
\end{figure}

To investigate that subgraphs indeed produce superior performance, we experimented with a single big graph that includes all training samples simultaneously. When compared with semi-supervised audio node classification task on IEMOCAP, the large graph achieves only a 49.44\% recognition accuracy which is much lower than the accuracy (63.84\%) achieved using the subgraphs-based learning framework, both without SSL. An intuitive explanation of the superior performance of subgraphs for audio classification: Since the graph structure is not given, constructing smaller graphs with balanced class samples introduces less error (as compared to a large graph) as subgraphs limit the number of unlabeled nodes per graph. 
\begin{figure}[t]
  \centering
\includegraphics[width=1\linewidth, trim=0.26cm 0mm 5.6cm 25mm, clip=true, bb=0 0 936 612]{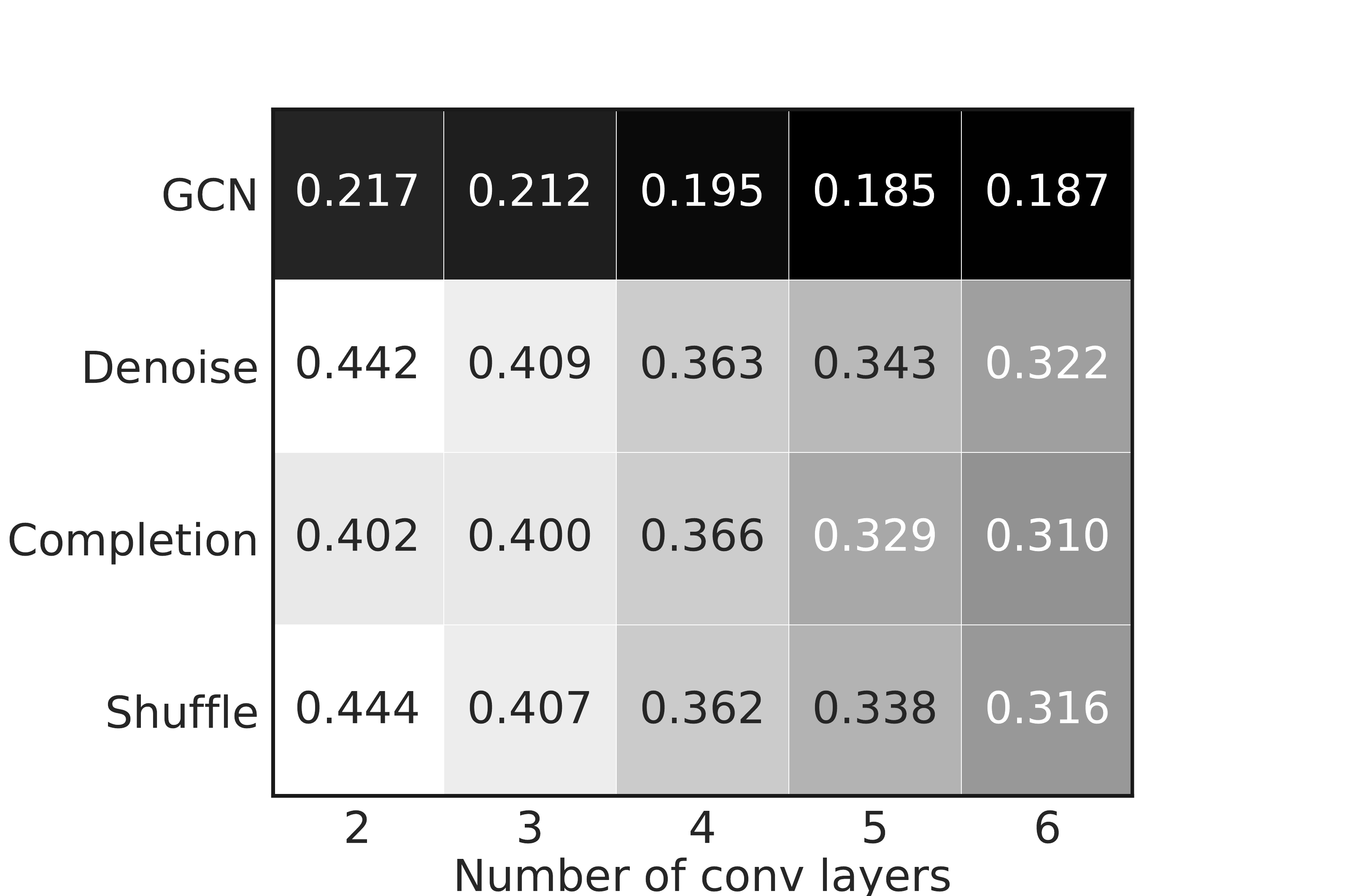}
  \caption{Oversmoothness analysis for IEMOCAP dataset: Measured in terms of mean average distance (MAD), is significantly smaller for our SSL-based models compared to GCN models. Darker color indicates smaller MAD values i.e., higher oversmoothing. We observed a similar trend for other datasets studied in our work.}\label{fig:MAD}
  \vspace{-2mm}
\end{figure}
Fig.~\ref{fig:Hyper_param_analysis}(a) shows how the classification performance varies with graph size (number of nodes). We observe that initially the recognition accuracy improves as the graph size increases (up to size $50$), but then starts decreasing. Overall, our observation is that subgraphs are more effective than using larger graphs in settings where we have limited labeled and a large amount of unlabeled training data. 
\subsubsection{Number of masked nodes in SSL tasks} 
For the completion and the shuffle SSL tasks, we select a fraction of nodes (masked nodes) within a subgraph to apply the transformation. Fig.~\ref{fig:Hyper_param_analysis}(b) shows how the  fraction of masked nodes affects the classification performance. We observe that in general, the recognition accuracy drops as masked nodes are increased beyond 10\%.
\begin{table}[t]
\centering
\caption{Model robustness against noise: Performance changes in recognition accuracy (in \%) for noisy speech vs. clean speech. The results shown are for the IEMOCAP database where $\downarrow$ indicates a drop in performance.
}
\renewcommand*{\arraystretch}{1.2}
\resizebox{1.0\linewidth}{!}{
\small
\begin{tabular}{l |ccccr}
\toprule
 & \multicolumn{5}{c}{\emph{Speech noise types}}\\
 \cline{2-6}
& Babble    & Factory   & White & Pink  & Cockpit   
\\ \midrule
Ours w/o SSL      & $2.8 \downarrow$ & $1.6\downarrow$ & $3.1\downarrow$ & $2.0\downarrow$ & $1.1\downarrow$ 
\\ 
Ours w/ denoise   & $\mathbf{0.5}\downarrow$ & $1.2\downarrow$ & $\mathbf{0.5}\downarrow$ & $\mathbf{0.9}\downarrow$ & $\mathbf{0.8}\downarrow$  
\\ 
Ours w/ completion    & $0.7\downarrow$ & $1.1\downarrow$ & $0.8\downarrow$ & $1.3\downarrow$ & $1.6\downarrow$ 
\\ 
Ours w/ shuffle      & $0.8\downarrow$ & $\mathbf{0.8}\downarrow$ & $1.0\downarrow$ & $1.5\downarrow$ & $0.9\downarrow$  
\\ 
\bottomrule
\end{tabular}
}
\label{tab:robust}
\vspace{-3mm}
\end{table}

\subsubsection{Oversmoothness} GCN models are known to suffer from oversmoothness as the number of layers increases \cite{chen2020measuring}. We investigate whether our model benefits from SSL in mitigating oversmoothness. A quantitative metric for measuring oversmoothness is Mean Average Distance (MAD). It measures the average distance from a node to all other nodes \cite{chen2020measuring}. Using MAD, we experimented with a varying number of graph convolution layers on the IEMOCAP database. Fig.~\ref{fig:MAD} shows that SSL-based models are more resilient to oversmoothness, producing more distinguishable node embeddings. This is clear from the average distances being much larger compared to the no SSL case. A similar trend is observed in other datasets.

\subsubsection{Robustness against noise} To investigate the robustness of our model against noisy data, we experiment with six common noise types that may corrupt speech data: babble, factory, white, pink and cockpit noise \cite{kumar2019dirichlet}. Assuming additive noise, the noisy mixtures are obtained by adding the noise (one type at a time) to each speech sample first and then using the feature encoders as usual. Noise is added only to the test samples during inference. We compare the performance of our model on noisy test data with that with clean test input in Table \ref{tab:robust} for the IEMOCAP dataset. Clearly, SSL provides significant robustness against noise as their drop in performance is consistently smaller compared to GCN semi-supervised results without SSL.
\begin{figure}[t]
    \centering
    \includegraphics[width=0.8\linewidth, trim=1mm 0mm 0mm 0mm, clip=true, bb={0 0 617 423}]{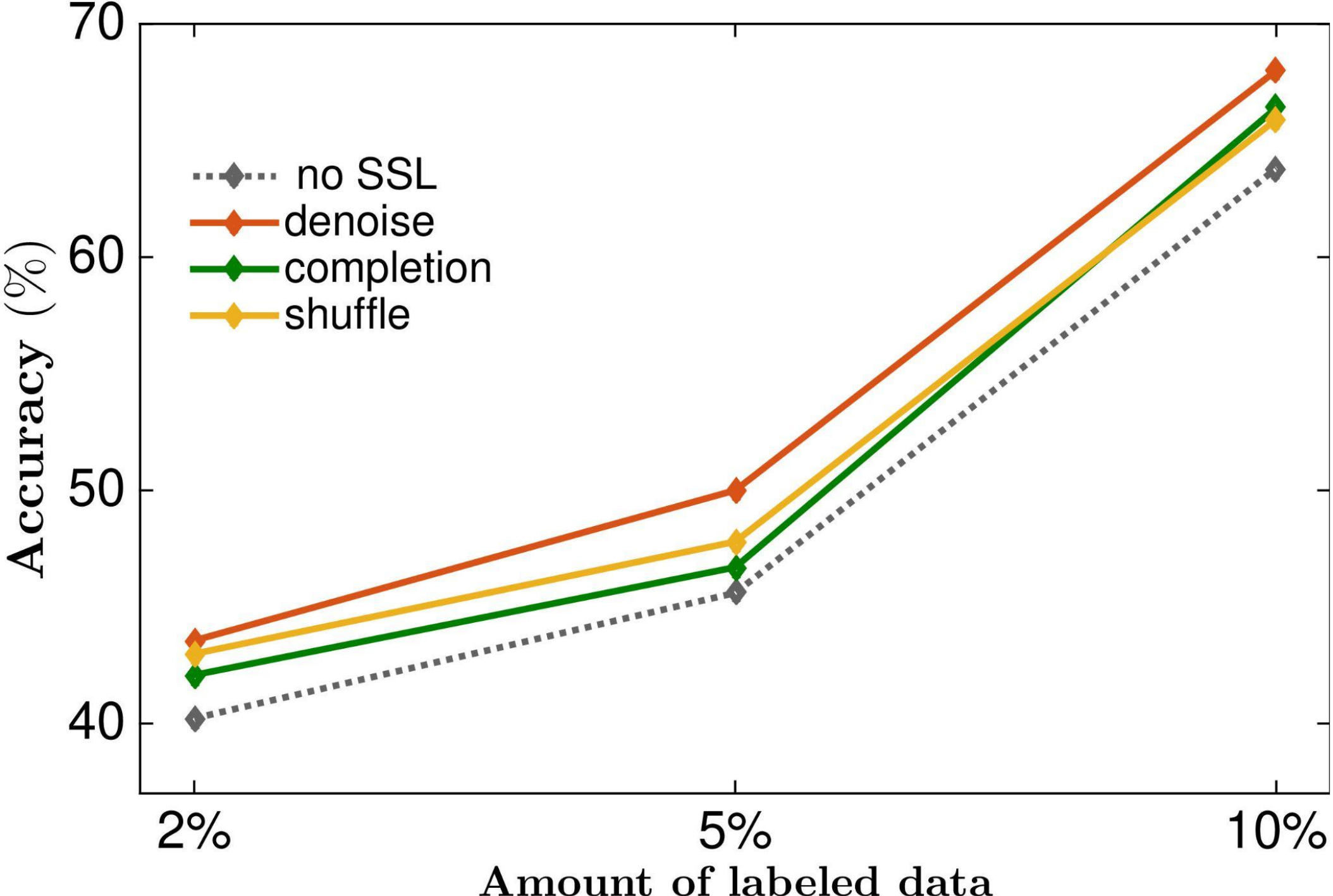}
    \caption{Performance using less than 10\% labeled data: Our model performs reliably with SSL producing consistent improvement as labeled data becomes scarce; the denoise task performs the best (results shown for IEMOCAP).}
    \label{fig:2510per}
    \vspace{-3mm}
\end{figure}
\subsubsection{Reducing labelled data further} We further investigate how the subgraph strategy holds up to even more scarce labelled data, and whether or not the SSL tasks help. Fig.~\ref{fig:2510per} shows the performance of our model with 2\% and 5\% labeled data in addition to 10\% on the IEMOCAP database. We note that SSL tasks bring consistent improvement across all cases, where the denoise task performs the best.
%
\section{Conclusion}
\label{sec:conclusions}
Our work contributes to the understanding of semi-supervised audio representation learning  - a relatively understudied topic in the acoustics and speech community. We developed a subgraph-based SSL framework for audio representation learning with limited labelled data. We make use of graphs to capture the underlying information in the unlabeled training samples and their relationship with the labelled samples. To this end, we proposed an effective subgraph construction technique and a new graph SSL task (graph shuffling). Our framework is generic, and can effectively handle both speech and non-speech audio data. Our model could achieve comparable or better performance than fully supervised models, despite using only 10\% of the labelled data. Since the graph structure in our task has to be constructed first, our model is currently not end-to-end learnable. This could be addressed in future work where the graph structure itself is learned jointly with the embeddings. Our current model relies on pre-trained embeddings, which gives the flexibility of choosing any suitable embeddings given a task. Nevertheless, our model can be made end-to-end trainable which will be addressed in a future work.
\par\textbf{Ethical/social impact statement:} Our work may be applied to classify speech and acoustics data. Therefore, it may be used to `hear' and should be used carefully. We used public databases that are mostly balanced in terms of male and female subjects. However, they are not balanced considering factors such as ethnicity, language spoken and other demographic factors. The speech emotion analysis application uses only four archetypal expressions. We are aware that human emotion is far more complex, and thus do not advocate the use of such systems for sensitive decision making areas. We also note that automatic emotion classification from speech is an evolving area of research and questions of fairness and universality remain to be explored.

\bibliographystyle{unsrt}
\bibliography{biblo}

\begin{IEEEbiography}[{\includegraphics[width=1in,height=1.25in,clip,keepaspectratio]{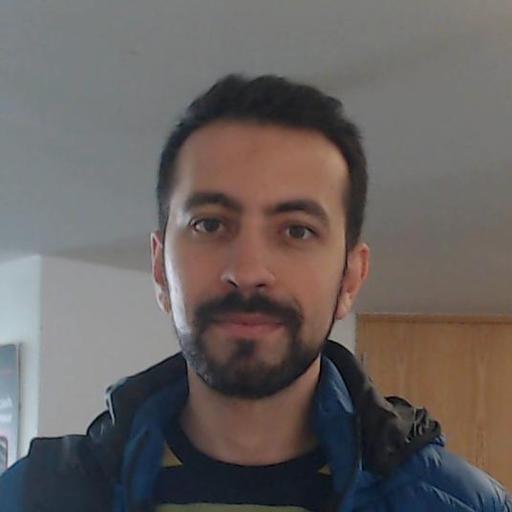}}]{Amir Shirian} is currently a PhD student in the Department of Computer Science at the University of Warwick, UK. He has received his BSc (2015) and MSc (2018) degrees in Electrical Engineering from the University of Tehran, Iran. His research interests include multimodal signal processing, graph neural networks, and machine learning with applications to emotion and behaviour understanding. 
\end{IEEEbiography}
\vspace{-35mm}
\begin{IEEEbiography}[{\includegraphics[width=1in,height=1.25in,clip,keepaspectratio]{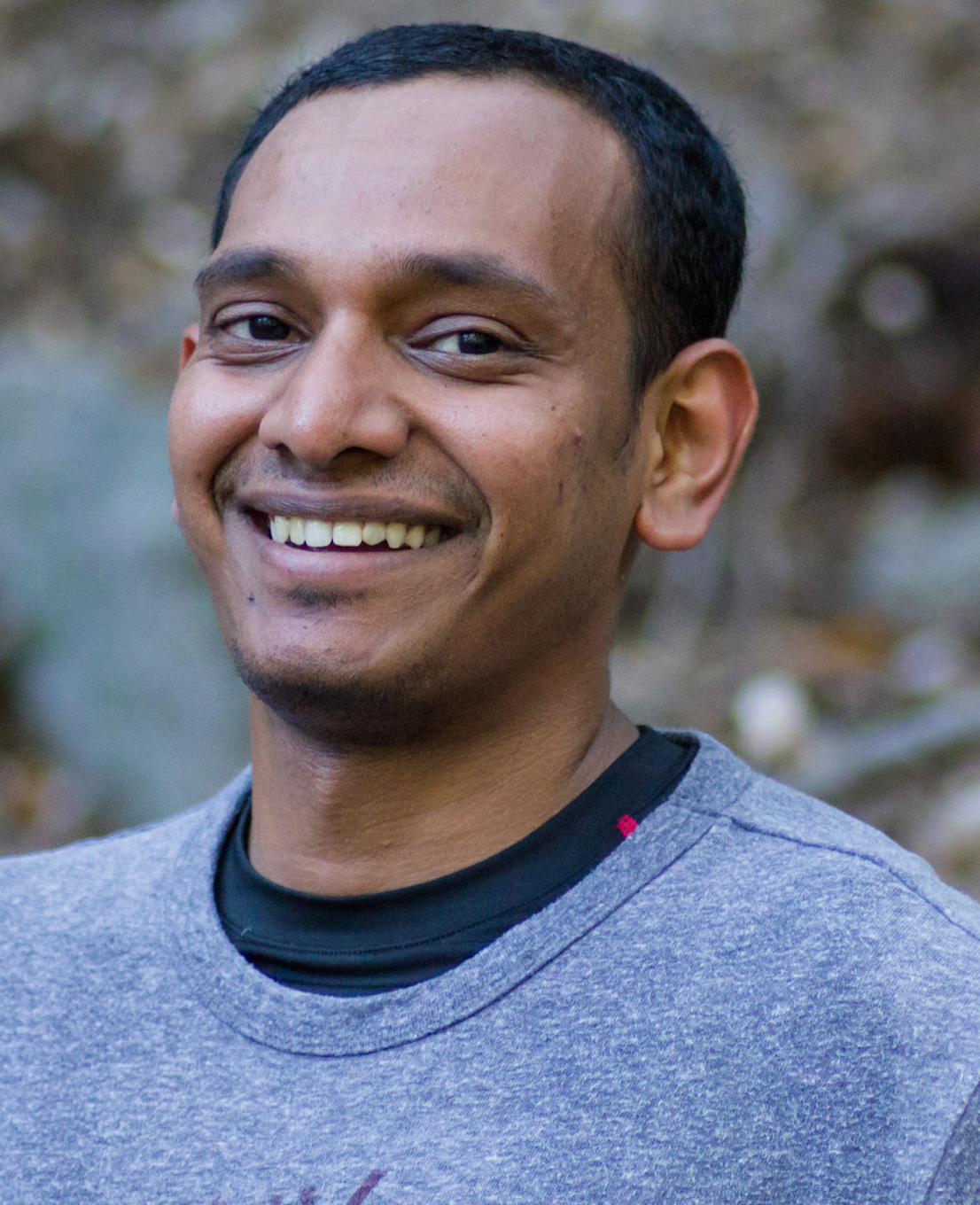}}]{Krishna Somandepalli }received his PhD in Electrical and Computer Engineering from the University of Southern California, CA, USA and a Masters degree from the University of California at Santa Barbara, CA, USA in Electrical and Computer Engineering. Following his Masters degree, he worked as an assistant research scientist at NYU Langone medical Center, New York, NY, USA. He currently works at Google Research.
His research interests include multimodal machine learning, media understanding and developing inclusive technologies.
\end{IEEEbiography}
\vspace{-35mm}
\begin{IEEEbiography}[{\includegraphics[width=1in,height=1.25in,clip,keepaspectratio]{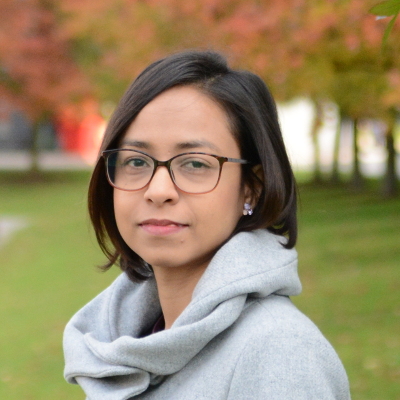}}]{Tanaya Guha} has received her PhD in Electrical and Computer Engineering from the University of British Columbia, Vancouver, Canada. She is currently a Senior Lecturer (Associate Professor) in the School of Computing Science, University of Glasgow, UK. Her research is focused on modelling and analysis of audio and visual data combining machine learning and signal processing with applications in healthcare and autonomous systems. She is a member of the Editorial Boards of Nature Scientific Reports and APSIPA Transactions on Signal and Information Processing. She is an elected member of IEEE Multimedia Systems Applications Technical Committee. She regularly serves in the organizing and program committees of ICME, ICMI, WACV and INTERSPEECH.
\end{IEEEbiography}

\end{document}